\documentclass{article}

\usepackage[preprint]{corl_2026} 

\usepackage{graphicx}
\usepackage{amsfonts}
\usepackage{wrapfig}
\usepackage{booktabs}
\usepackage{enumitem}
\usepackage[capitalise]{cleveref}
\graphicspath{{fig/}}

\makeatletter
\DeclareRobustCommand\onedot{\futurelet\@let@token\@onedot}
\def\@onedot{\ifx\@let@token.\else.\null\fi\xspace}

\makeatother

\newcommand{\method}{SEED-UMI}

\title{\method: Sharing the Exoskeleton between human and robot for onE-to-one Dexterous demonstration}

\author{
  Tengbo Yu \textsuperscript{\rm 1,2,}\thanks{Indicates equal contribution.} \quad
  Jiahao Wu \textsuperscript{\rm 2,}\footnotemark[1] \quad
  Daohan Li \textsuperscript{\rm 2,}\footnotemark[1] \quad
  Bingxu Chen \textsuperscript{\rm 2} \quad \\
  \bfseries Hao Liu \textsuperscript{\rm 2} \quad
  Xiaojian Ma \textsuperscript{\rm 2} \quad
  Hangxin Liu \textsuperscript{\rm 1,}\thanks{Corresponding author. Email: \texttt{hx.liu@pku.edu.cn}} \quad \\
  \textsuperscript{\rm 1} State Key Laboratory of General Artificial Intelligence, \\ School of Intelligence Science and Technology, Peking University \\
  \textsuperscript{\rm 2} Delta Intelligence \\
  \url{https://tengbo-yu.github.io/SEED-UMI/}
}

\begin{document}
\maketitle

\begin{abstract}
    Imitation learning for dexterous hands is bottlenecked by the difficulty of collecting contact-rich demonstrations that transfer faithfully to the robot. Prior wearable-exoskeleton systems record only on the human side and retarget via open-loop mappings calibrated in free space, which degrade under contact. We present \textbf{\method{}}, a framework in which \emph{both the human and the robot wear the same exoskeleton}: joint encoders become a physically shared measurement, and wrist cameras mounted to the exoskeleton observe the same outer mechanism during both human data collection and robot policy rollouts. This turns retargeting into paired cross-embodiment supervision and lets policies train directly on raw exoskeleton-centric wrist images, without segmentation or inpainting. On five contact-rich tasks, \method{} achieves $3.0\times$ greater data collection efficiency than exoskeleton-based teleoperation and a 70.0\% average rollout success rate.
\end{abstract}


\section{Introduction}
\label{sec:intro}


Dexterous hand manipulation remains one of the most challenging problems in robot learning due to the high degrees of freedom of robotic hands and the complex contact-rich interactions involved during manipulation. Although imitation learning (IL) has enabled robots to acquire sophisticated locomotion and arm manipulation skills, progress in dexterous manipulation has been comparatively limited. A major bottleneck lies in demonstration collection and retargeting~\cite{argall2009survey,osa2018algorithmic,rajeswaran2017learning,arunachalam2023dexterous}: accurately capturing natural human hand motions while transferring them reliably to robot remains difficult because human and robot embodiments differ in kinematics, sensing, and contact behavior~\cite{meattini2022human}.

Existing demonstration collection approaches each involve a trade-off between accurately recording dexterous hand motions and preserving natural hand behavior, which are often competing. Vision-based hand tracking allows demonstrations to be collected without constraining hand motion~\cite{du2012markerless,handa2020dexpilot,qin2023anyteleop,wang2024dexcap}, but data quality would degrade due to occlusion and temporal jitter when contacting with external objects. Glove-based systems equipped with inertial or flex sensors~\cite{hammond2014toward,liu2017glove,liu2019high,manus} provide more reliable motion data for interactive tasks, but are prone to skin-motion artifacts and require complex correspondence mapping between human and robot joints. Exoskeleton-based devices produce the most consistent hand gesture measurements that are identical to mechanically constrained hand motions in the robot side, yet often sacrifice operator comfort and demonstration throughput due to bulky and restrictive hardware~\cite{ben2014sensing,zhang2025doglove}.

More importantly, these collection approaches fundamentally shape the downstream learning problem. Vision-based systems typically simplify dexterous manipulation into end-effector control because fine-grained finger motions are more difficult to recover reliably~\cite{ding2025bunny}. Glove-based methods require extensive post-processing and retargeting to establish correspondence with robot hand kinematics~\cite{xin2026analyzing,liu2024reconfigurable}. As a result, current dexterous imitation learning pipelines largely treat human demonstrations and robot execution as two separate domains connected through increasingly complex retargeting models.

\begin{figure}[t]
    \centering
    \includegraphics[width=\linewidth]{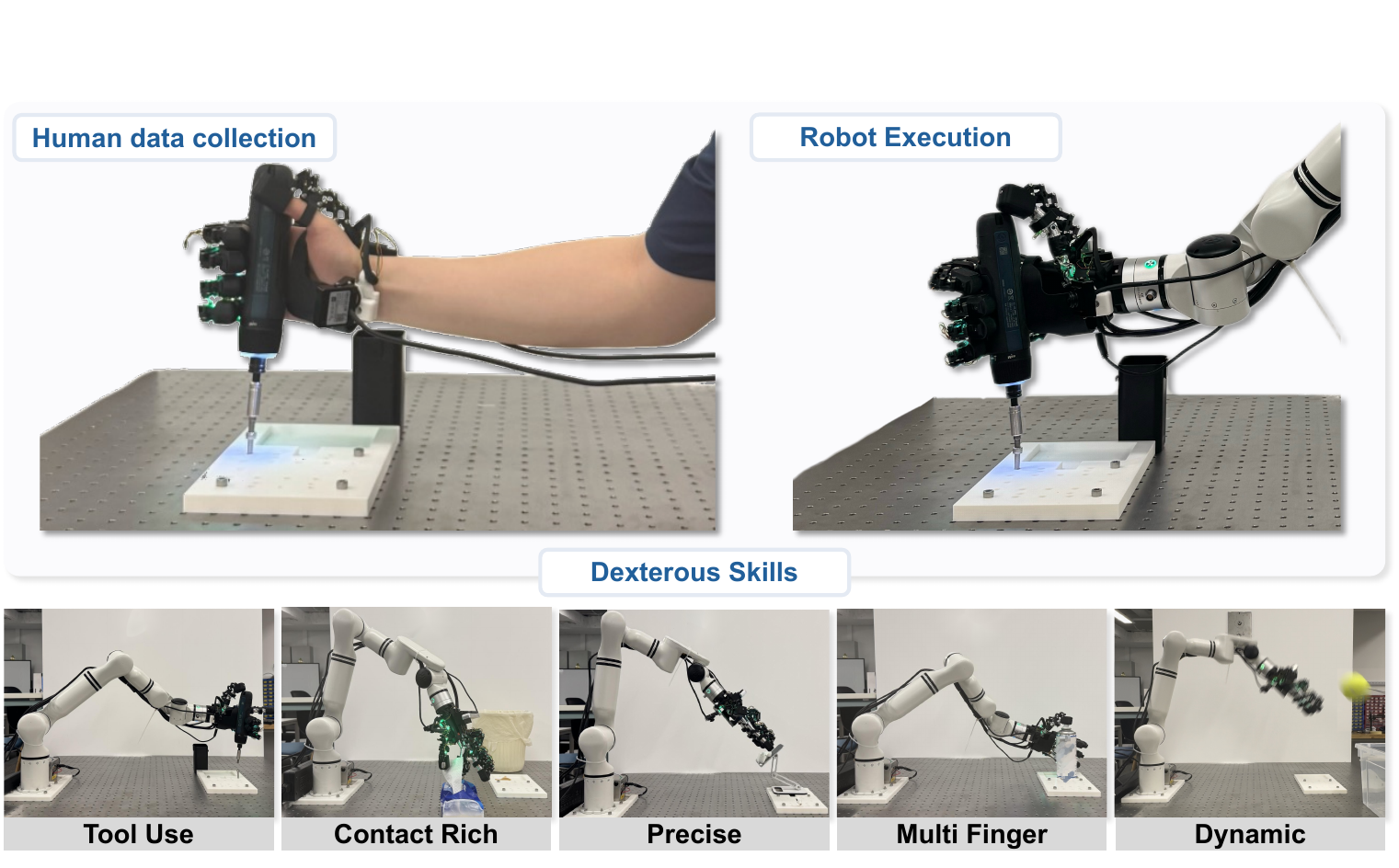}
    \caption{\textbf{\method{} data-collection interface and task suite.} \emph{Top:} the human operator and the dexterous robot hand wear the same exoskeleton, so demonstrations and robot replay are recorded through a shared physical device with matched encoders and wrist-mounted cameras. \emph{Bottom:} representative contact-rich tasks enabled by the interface, spanning tool use, small-object placement, grasp-and-throw timing, contact-rich table cleaning, and multi-finger spray actuation.}
    \label{fig:teaser}
    \vspace{-1.0em}
\end{figure}

Recently, exoskeleton-based systems have attracted renewed interest because they offer a promising way to reduce this correspondence gap through co-design with the target robot hand~\cite{dexumi,mirage,roviaug}. By aligning joint encoders with robot degrees of freedom and mounting wrist cameras from the robot end-effector viewpoints, such systems enable learning of more dynamic and reactive manipulation behaviors. However, existing approaches still operate as one-way mappings from human motion to robot motion. They typically calibrate retargeting in free space~\cite{dexumi}, which often degrades under contact-rich interactions due to friction, external loads, and joint coupling effects. Furthermore, systems such as DexUMI~\cite{dexumi,mirage,roviaug} rely on generative image translation to bridge visual discrepancies between human and robot observations, while robot-side measurements themselves are not used to improve correspondence learning.

In this paper, we present \textbf{\method{}}, a demonstration-collection and policy-learning framework built upon the idea of \emph{shared physical measurement}. Instead of learning a mapping between separate human and robot hand embodiments, we design a shared measurement interface that can be directly used by both. Specifically, we develop a kinematically isomorphic exoskeleton co-designed with the target robot hand, whose link lengths, joint axes, and motion ranges are jointly optimized such that the same exoskeleton device can be worn by both the human hand and the robot hand. As a result, both human demonstrations and robot executions generate encoder measurements under the same mechanical structure across all 20 degrees of freedom (\cref{sec:hardware}).

Building upon this shared interface, we learn a contact-aware encoder-to-command mapping through a two-stage process consisting of motor babbling followed by paired trajectory fine-tuning (\cref{sec:software}). Importantly, discrepancies between human and robot encoder trajectories under the same intended motion naturally become paired supervisory signals for retargeting and policy learning. In addition, policies are trained directly from raw exoskeleton-centric wrist observations, avoiding post-hoc segmentation and generative image translation pipelines.

Combining these components, \method{} enables natural, contact-rich demonstrations collected from human operators to be transferred directly to robot execution, decouples human demonstration collection from real-time robot teleoperation, simulation, or reinforcement learning. We evaluate our framework on five dexterous manipulation tasks spanning tool use, precise object placement, grasp-and-throw timing, contact-rich table cleaning, and multi-finger spray actuation (\cref{sec:evaluation}). \method{} achieves a data collection speed 3.0$\times$ faster than exoskeleton-based teleoperation. With paired fine-tuning, learned policies achieve a mean success rate of 70.0\% across all tasks, approaching the 71.7\% teleoperation baseline while outperforming it on highly precise and dynamic tasks such as throwing.

Our results suggest that overcoming the demonstration collection and retargeting bottleneck in dexterous manipulation may require rethinking the interface between human and robot embodiments rather than solely improving retargeting methods. By \textit{transforming retargeting and visuomotor policy learning into a form of paired cross-embodiment supervision}, our idea of shared physical measurement offers a promising direction toward more scalable and efficient dexterous robot learning.

\section{Related Work}
\label{sec:related}

\begin{figure}[t]
    \centering
    \includegraphics[width=\linewidth]{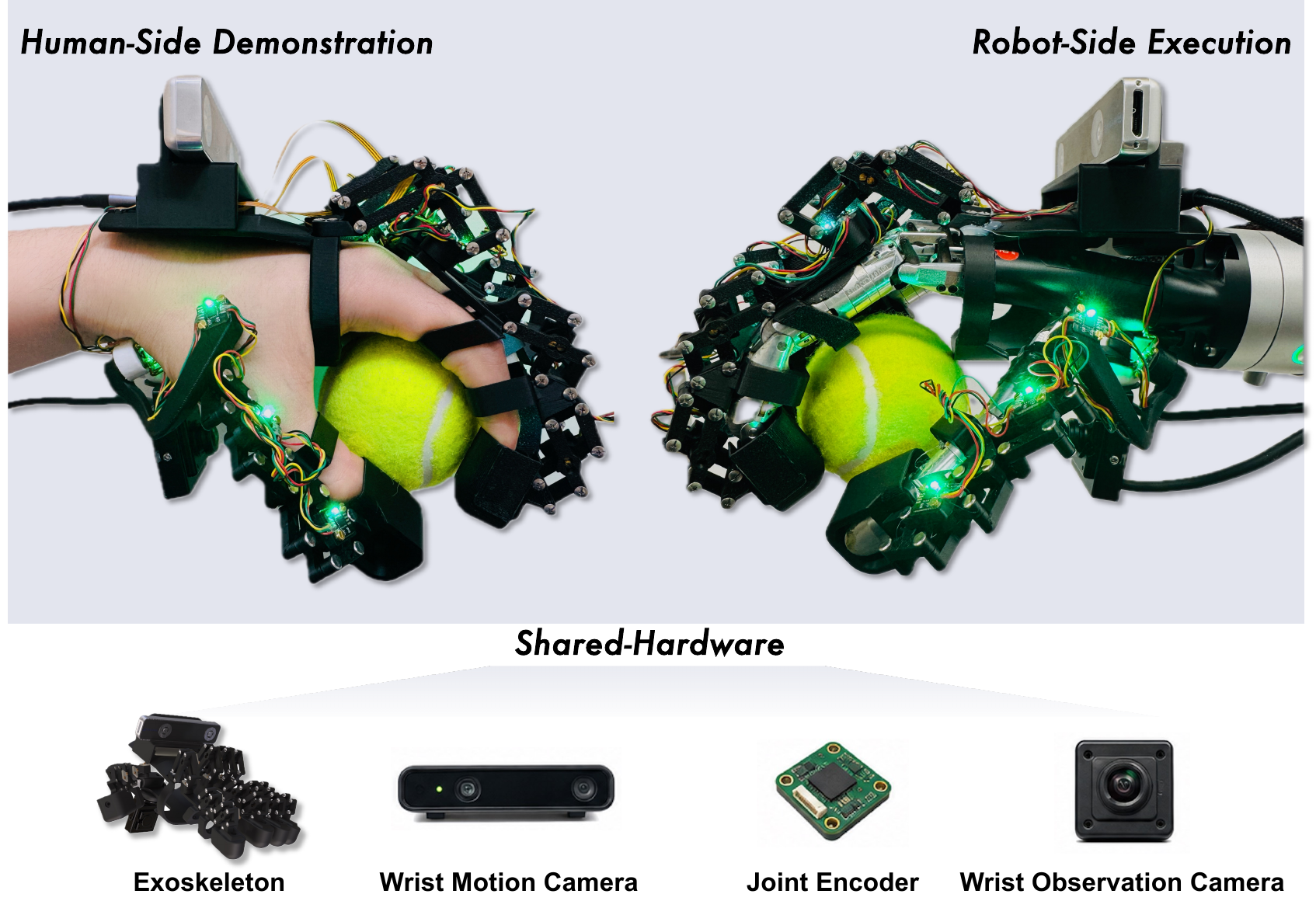}
    \caption{\textbf{Shared exoskeleton hardware.} The same skeleton is worn by the human operator (\emph{left}) and the fully-actuated target hand (\emph{right}). One contactless rotary encoder per joint, a dorsal T265 for 6-DoF wrist tracking, and a ventral fisheye camera are rigidly mounted on the exoskeleton frame, so the encoder vector $\mathbf{e}_t$ and the wrist-camera intrinsics are identical across embodiments.}
    \label{fig:hardware}
    \vspace{-1.0em}
\end{figure}

\paragraph{Wearable exoskeletons for dexterous demonstration.}
Glove-based systems~\cite{chi2024universal,manus,liu2017glove,liu2019high,hammond2014toward,liu2024reconfigurable,zhang2025doglove,hato} record human hand pose but lack one-to-one joint correspondence with the target robot. Wearable~\cite{dexumi,whed,dexop,mile,dexvitac} constrain the hand into a robot-matched kinematic structure: DexUMI~\cite{dexumi} instruments the exoskeleton with joint encoders and adapts it to multiple robot hands; WHED~\cite{whed} improves wearability with a pose-tolerant thumb mechanism. These systems mount the exoskeleton on the human side only and transfer demonstrations via a one-shot free-space regression. \method{} extends this paradigm by mounting the same exoskeleton on the robot, converting the open-loop mapping into a supervised learning problem with paired cross-embodiment data.

\paragraph{Visual alignment for cross-embodiment policy learning.}
Inpainting-based methods~\cite{mirage,roviaug} and the visual pipeline of DexUMI~\cite{dexumi} bridge the appearance gap between embodiments by hallucinating the target hand in the observation stream. These operate at the policy interface and require generative models. We instead remove most of the appearance gap by design: because both embodiments wear the same exoskeleton and the wrist camera is rigidly mounted to that exoskeleton, data collection and policy rollouts expose the policy to the same outer mechanism, object, and contact interface. \method{} therefore trains on raw wrist images with standard visual augmentation, without hand segmentation or generation.

\paragraph{Encoder-to-motor mapping.}
DexUMI~\cite{dexumi} fits a per-joint regression in free space, which degrades under contact load on fully-actuated hands. DexterityGen~\cite{dexteritygen,rajeswaran2017learning,gupta2016softdexterous,andrychowicz2020inhand,openai2019rubik,dexmimicgen} trains a low-level controller via large-scale simulation RL. Our approach occupies a middle ground: we obtain real-world supervision by replaying demonstrations on the robot through the shared exoskeleton, requiring neither simulation nor reinforcement learning, in the spirit of demonstration-augmentation methods~\cite{mimicgen,demogen,dexmv,ilad} but specialized to the encoder-to-command mapping.

\section{Hardware: A Shared Exoskeleton}
\label{sec:hardware}

\begin{figure*}[t]
    \centering
    \includegraphics[width=\linewidth]{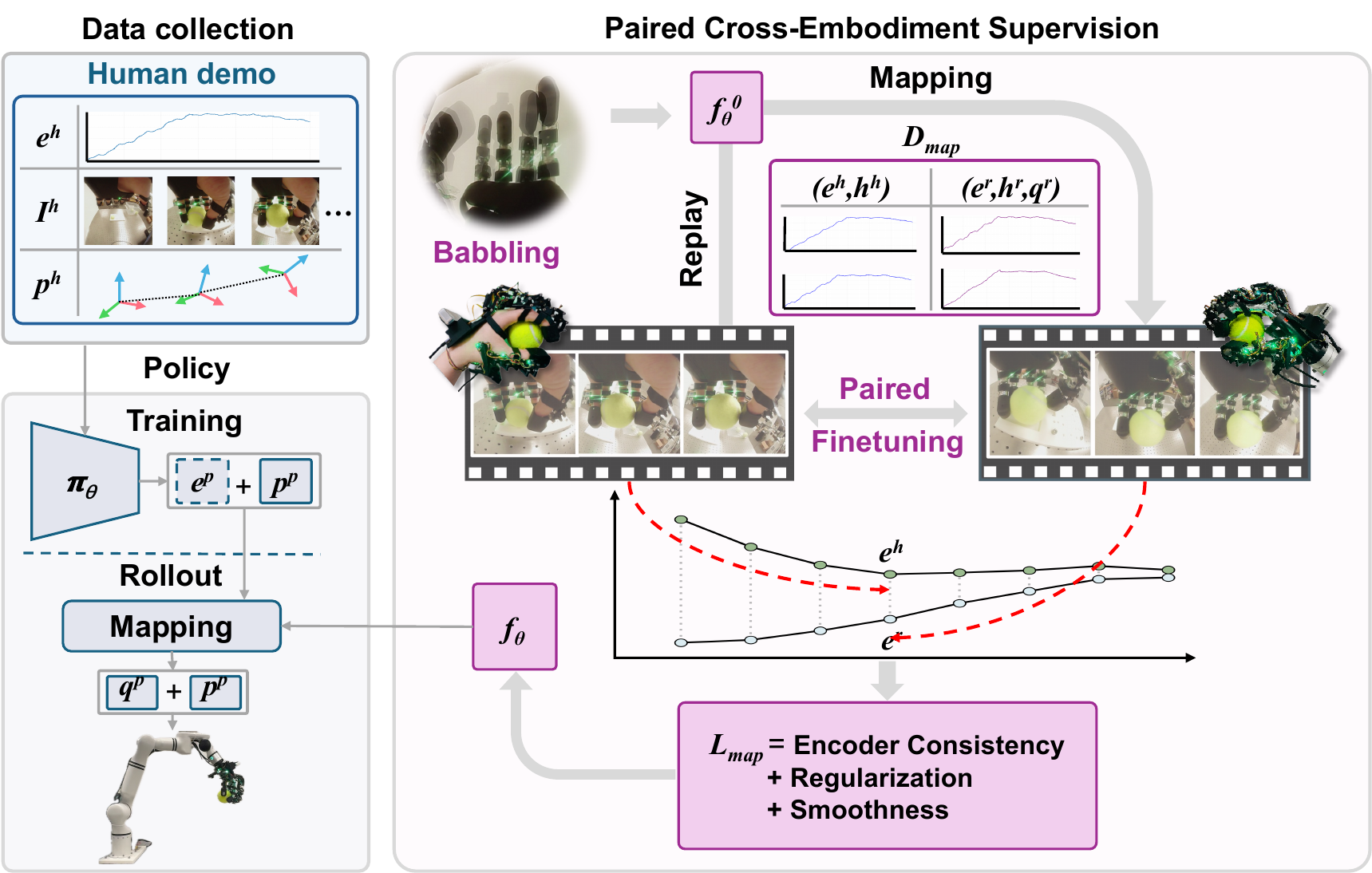}
    \caption{\textbf{\method \ pipeline.} Human demonstrations are collected through the shared exoskeleton, paired robot replay refines the encoder-to-command mapping, and the
  learned policy rolls out on the robot with matched exoskeleton-centric observations.}
    \label{fig:pipeline}
    \vspace{-1em}
\end{figure*}


The hardware of \method{} matches the contact-facing exoskeleton geometry and encoder frame on the human and robot sides, while adapting cuffs and mounts to each embodiment. This makes encoder readings a shared \emph{cross-embodiment measurement}: discrepancies under the same command supervise the residual command-to-joint mapping (\S\ref{sec:software}). Figure~\ref{fig:hardware} shows both systems.


\subsection{Exoskeleton Mechanism Design}
\label{sec:hardware:mechanism}

The exoskeleton is co-designed with a single fully-actuated dexterous hand as its sole target platform. Because we do not pursue cross-platform transferability, we can fix link lengths, joint axes, and range-of-motion limits to a direct mechanical replica of the robot's kinematic tree, so that we can enforce two stronger properties: 1:1 joint correspondence on \emph{both} sides of the device, and a contact geometry that is identical between the human-worn and robot-worn instances.

\paragraph{E.1 Isomorphic kinematic skeleton.}
The skeleton has one revolute joint per active degree of freedom of the target hand, with link lengths matched to the robot's phalanx lengths and joint axes oriented identically. Hard stops mirror the robot's range-of-motion limits, so configurations the operator can reach are exactly those the robot can reach. The exoskeleton instantiates 20 independently measured joints across five fingers.

\paragraph{E.2 Parallel four-bar linkage for encoder routing.}
\begin{wrapfigure}{r}{0.5\linewidth}
    \vspace{-1.4em}
    \centering
    \includegraphics[width=\linewidth]{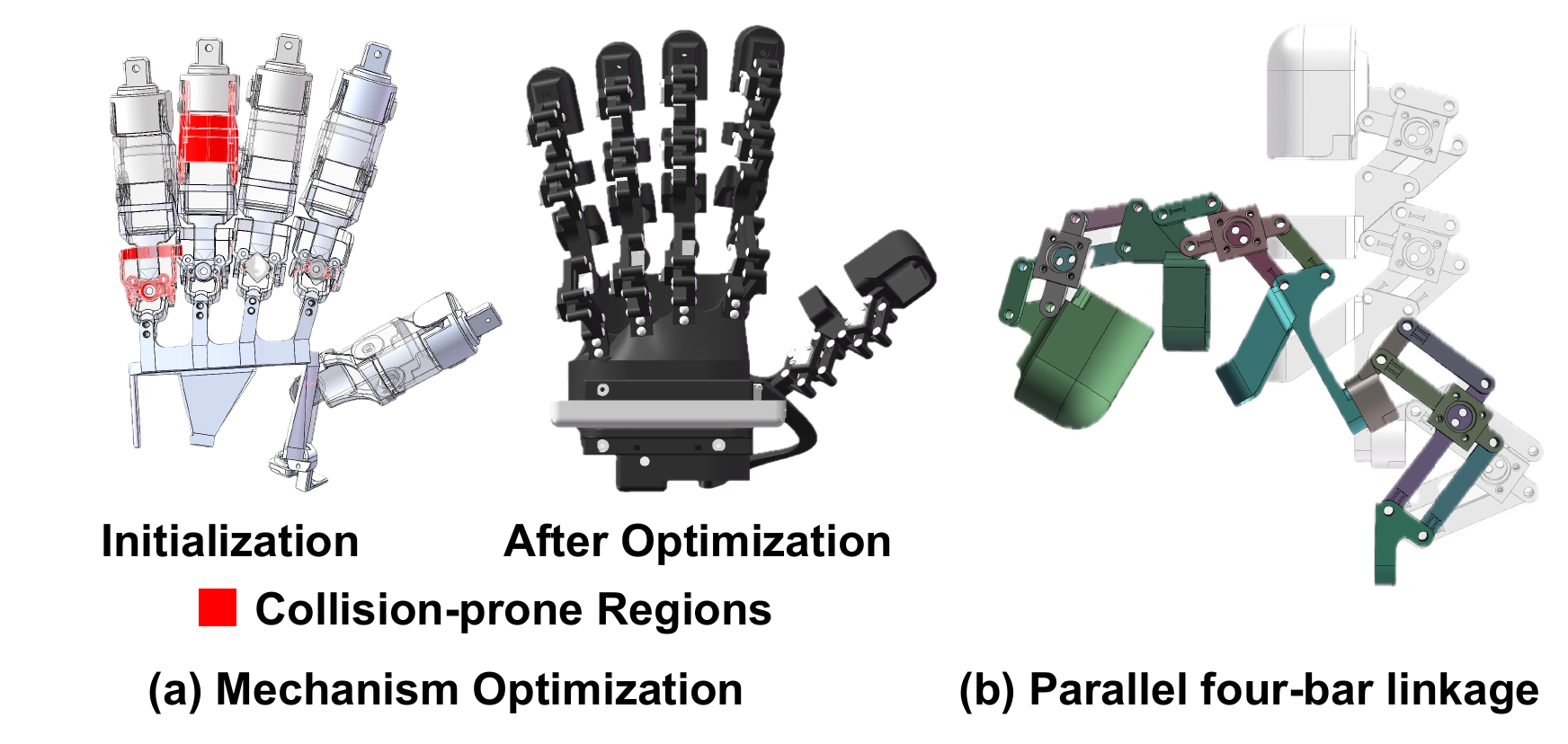}
    \caption{\textbf{Hardware Design.} (a) shows the collision avoidance of our design during operation. (b) shows how the linkage work during operation.}
    \label{fig:design}
    \vspace{-1.4em}
\end{wrapfigure}
A naive layout that places the encoder directly on the lateral interphalangeal axis is mechanically simple but fails two requirements at once: the housing protrudes into the lateral clearance envelope and restricts abduction\allowbreak/\allowbreak adduction below the robot's native range; it also occludes part of the palmar contact surface, breaking the shared-geometry property between embodiments. We therefore route the rotation through a \emph{parallel four-bar linkage} to a dorsally mounted contactless magnetic encoder (Figure~\ref{fig:design} (b)). The linkage preserves one-to-one angular correspondence between the joint axis and encoder shaft, while the palmar and lateral surfaces remain unobstructed (Figure~\ref{fig:design} (a)). Link lengths are tuned per finger to minimize dorsal height, and press-fit magnet pockets with precision sleeve bearings at each pivot keep encoder alignment stable across repeated wear cycles.

\paragraph{E.3 Identical contact geometry on both sides.}


The human- and robot-worn instances share the same outer skeleton: link lengths, joint axes, dorsal linkage, and palmar fingertip pads. Cuffs and mounts adapt this skeleton to each embodiment without changing joint frames. Thus both embodiments present the same external contact geometry to the object, and human--robot encoder discrepancies mainly reflect load-dependent effects such as contact force, friction, and hysteresis, which provide the residual signal used in \S\ref{sec:software}.

\subsection{Sensor Integration}
\label{sec:hardware:sensors}

All sensing is mounted on the exoskeleton frame so that every signal is recorded in the same physical frame regardless of which embodiment wears the device. Two design objectives drive the sensor layout: (i)~capture the action and observation channels needed for downstream policy learning, and (ii)~minimize the distribution shift of those channels between the human-side and robot-side.

\paragraph{S.1 Joint encoders.}
To precisely capture joint actions, our exoskeleton integrates contactless magnetic rotary encoders at every actuated joint on \emph{both} embodiments---the human-worn and the robot-worn instances share the same encoder placement. Concatenating per-joint readings yields $\mathbf{e}_t \in \mathbb{R}^{20}$. Due to the four-bar linkage routing and manufacturing tolerances, the mapping between exoskeleton encoder readings $\mathbf{e}_t$ and robot joint commands $\mathbf{q}_t$ is non-linear; we therefore learn this mapping with a data-driven model described in~\S\ref{sec:software}.

\paragraph{S.2 Wrist-mounted cameras.}

A dorsal Intel RealSense T265 records 6-DoF wrist pose, while a ventral fisheye camera ($150^{\circ}$ FoV) observes the workspace. Both are rigidly mounted to the exoskeleton, fixing their intrinsics and encoder-frame extrinsics across the human and robot sides. Because data collection and rollout use the same exoskeleton-camera assembly, the policy receives aligned raw observations $I_t$ with only standard photometric and crop augmentations. Unlike DexUMI~\cite{dexumi}, this hardware alignment requires no hand segmentation or inpainting, preserving contact pixels for fine manipulation.


\section{Software Adaptation: Two-Stage Cross-Embodiment Mapping}
\label{sec:software}



The shared exoskeleton gives both embodiments a common encoder signal $\mathbf{e}_t \in \mathbb{R}^{20}$, but robot execution still requires an encoder-to-command mapping $\mathbf{e}_t \mapsto \mathbf{q}_t$. Instead of fitting this mapping once in free space~\cite{dexumi}, we bootstrap it with robot motor babbling and refine it with paired replay under real contact. The mapping is proprioceptive, while aligned wrist images are reserved for policy learning; Figure~\ref{fig:pipeline} shows the full pipeline.

\paragraph{M.1 Motor babbling on the robot.}

The robot wears the exoskeleton and executes a structured exploration protocol, covering single-joint sweeps, coupled motions at varied speeds, and a few self-contact poses. We record robot joint positions $\mathbf{q}_t$ and the robot-side encoder reading $\mathbf{e}_t^{(r)}$, the history vector $\mathbf{h}_t$ concatenates the previous $K$ encoder readings, finite-difference encoder velocities, and a binary opening/closing direction indicator for each joint, and fit an initial mapping $f_{\theta^0}\colon (\mathbf{e}_t, \mathbf{h}_t) \mapsto \mathbf{q}_t$. This free-space model captures basic kinematics and history effects, but still drifts under contact, motivating the paired replay stage in M.2.

\begin{figure*}[t]
    \centering
    \includegraphics[width=0.9\linewidth]{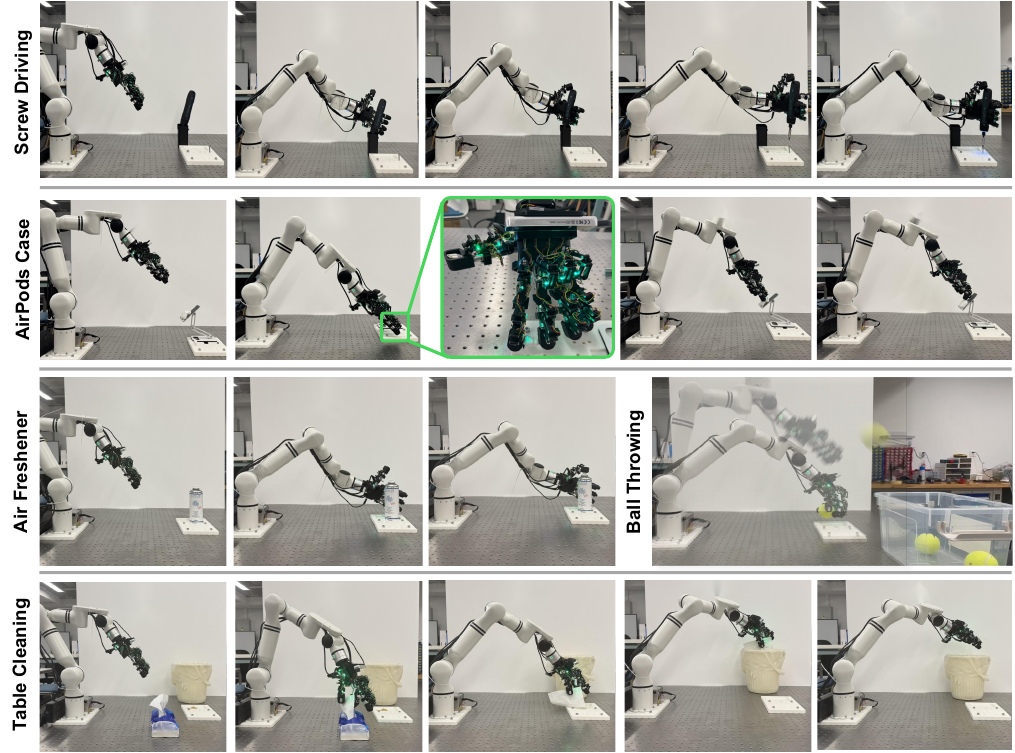}
    \caption{\textbf{Policy Rollouts.} We evaluate \method{} on five contact-rich dexterous manipulation tasks that together stress tool stabilization, index--middle finger coordination, grasp-to-release timing, contact-rich table wiping, and thumb-involved multi-finger actuation.}
    \label{fig:tasks}
    \vspace{-1.0em}
\end{figure*}

\paragraph{M.2 Paired replay and fine-tuning.}

The human operator performs representative object-grasping motions while wearing the exoskeletons, recording contact-rich encoder trajectories $\{\mathbf{e}_t^{(h)}, \mathbf{h}_t^{(h)}\}$. We replay each demonstration on the robot through $f_{\theta^0}$ and record $\{\mathbf{e}_t^{(r)}, \mathbf{h}_t^{(r)}, \mathbf{q}_t^{(r)}\}$ from robot-side exoskeleton. This produces a paired mapping dataset $\mathcal{D}_{\mathrm{map}} = \{(\mathbf{e}_t^{(h)}, \mathbf{h}_t^{(h)}, \mathbf{q}_t^{(r)}, \mathbf{e}_t^{(r)}, \mathbf{h}_t^{(r)})\}$, with a forward model $g_\phi$ from robot commands and encoder history to robot-side encoder readings, and fine-tune $f_\theta$ as an inverse problem:
\begin{equation}
    \mathcal{L}_{\mathrm{map}} =
    \| g_\phi(f_\theta(\mathbf{e}_t^{(h)}, \mathbf{h}_t^{(h)}), \mathbf{h}_t^{(h)}) - \mathbf{e}_t^{(h)} \|_2^2
    + \beta \| f_\theta(\mathbf{e}_t^{(h)}, \mathbf{h}_t^{(h)}) - \mathbf{q}_t^{(r)} \|_2^2
    + \lambda \| \Delta f_\theta \|_2^2,
    \label{eq:loss}
\end{equation}
The loss matches predicted robot-side encoders to the human trace, regularizes commands toward paired replay, and penalizes temporal jumps. We can repeat replay-and-refine once to turn the free-space babbling model into a contact-robust encoder-to-command function without manual retuning.

The paired replay stage does not assume that the initial replay perfectly reproduces the human trajectory. Instead, it uses the discrepancy between the human-side encoder trace and the robot-side encoder response as a correction signal for the command mapping.

\paragraph{M.3 Policy training and rollout.}
For downstream policy training we adopt the ACT~\cite{act}, Diffusion Policy (DP)~\cite{dp} and $\pi_{0.5}$~\cite{pi05}, taking the raw wrist observation $I_t$ and the encoder vector $\mathbf{e}_t$ as input and predicting an action chunk of length $L$ consisting of a 6-DoF wrist action and the 20-DoF hand command. All policy training is performed on human-collected real data; no simulation, reinforcement learning, hand segmentation, or generative inpainting is involved.

\begin{figure*}[t]
    \centering
    \includegraphics[width=0.9\linewidth]{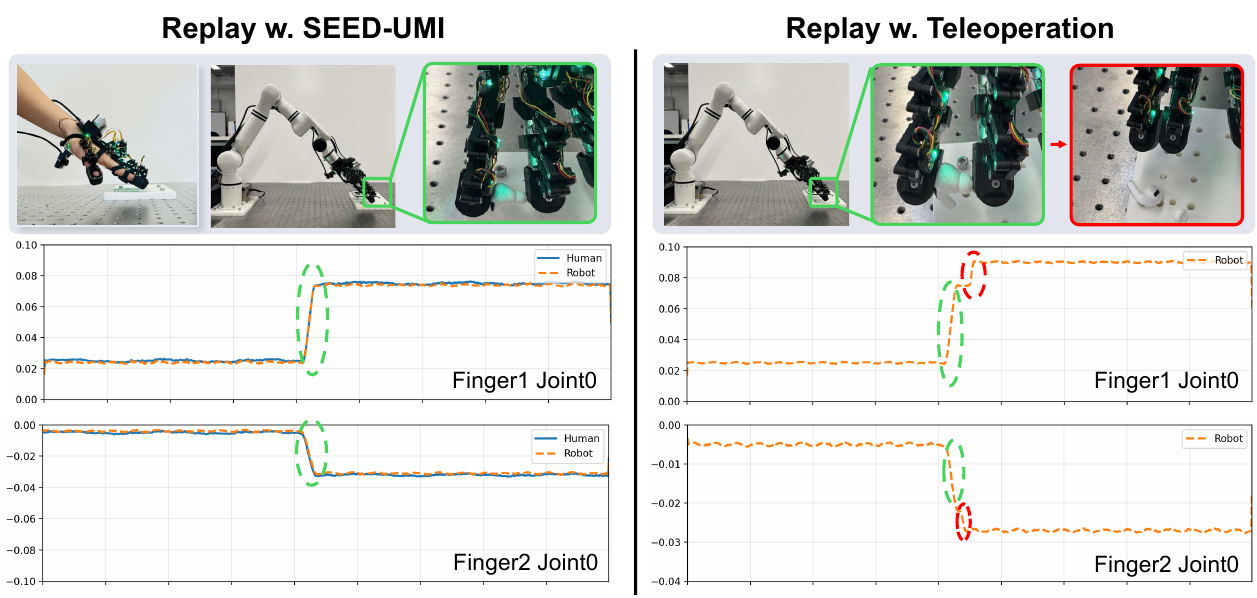}
    \caption{\textbf{Fine manipulation comparison.} AirPods task with human- (blue) and robot-side (orange) encoder traces.}
    \label{fig:comparisons}
    \vspace{-1.0em}
\end{figure*}

\paragraph {Qualitative Study}
This qualitative comparison highlights why the data-collection interface matters for fine manipulation (Figure \ref{fig:comparisons}). During \method{} collection, the operator directly performs AirPods insertion and can regulate delicate contacts through immediate visual and physical feedback. In contrast, teleoperation separates the operator from the object contact; the resulting replay can over-drive the motion and apply excessive force, which may damage fragile objects. This feedback advantage helps explain why \method{} is better suited to precise contact-rich demonstrations than teleoperation.
\section{Evaluation}
\label{sec:evaluation}

\subsection{Experiment Setup}
\label{sec:eval:setup}

\paragraph{Hardware.} Our platform consists of a RealMan RX75 robot arm and a Wuji dexterous hand with four actuated DoFs per finger (20 in total). The human and robot wear the shared exoskeleton described in \S\ref{sec:hardware}, preserving the outer contact geometry and camera placement across embodiments. A dorsal Intel RealSense T265 provides 6-DoF wrist pose, while a ventral fisheye camera with a $150^{\circ}$ field of view supplies policy observations. Wrist trajectories use a fixed T265-to-tool-center-point (TCP) alignment without trajectory correction.

\paragraph{Tasks.} We evaluate five real-world tasks (Fig.~\ref{fig:tasks}) covering tool use, fine finger coordination, dynamic release, and sustained contact:
\begin{itemize}[leftmargin=1.2em,itemsep=0.1em,topsep=0.2em,beginpenalty=10000]
    \item \textbf{Screw Driving:} grasp an electric screwdriver, align the bit with a screw head, and maintain axial pressure and a stable grasp during driving.
    \item \textbf{AirPods Case Insertion:} pick up an AirPod and insert it into its charging case using index--middle finger coordination; the case magnets assist final alignment.
    \item \textbf{Ball Basket Throwing:} grasp a ball from the table, lift it, and release it into a target basket, requiring accurate grasp-to-release timing.
    \item \textbf{Table Cleaning:} extract a tissue, wipe the tabletop under sustained contact, and discard the tissue.
    \item \textbf{Air Freshener Spray:} securely grasp a spray can and initiate a downward actuator press through coordinated multi-finger support and thumb motion.
\end{itemize}

\paragraph{Comparison.} We compare three conditions on the same robot platform, varying the demonstration source and encoder-to-command mapping:
\begin{itemize}[leftmargin=1.2em,itemsep=0.1em,topsep=0.2em,beginpenalty=10000]
    \item \textbf{Teleoperation (T).} The operator wears the exoskeleton, whose encoder readings are mapped directly to robot hand motor commands in real time. Policies are trained on demonstrations recorded on the robot side.
    \item \textbf{Ours w/o paired fine-tuning (P$^{-}$).} Policies are trained on human-side demonstrations and deployed using the babbling-only mapping $f_{\theta^0}$.
    \item \textbf{Ours w/ paired fine-tuning (P$^{+}$).} Policies are trained on human-side demonstrations and deployed using the refined mapping $f_\theta$ from the full \method{} pipeline (Fig.~\ref{fig:pipeline}).
\end{itemize}

\paragraph{Training data and mapping calibration.} Each policy is trained on 100 demonstration episodes per task, with initial object positions randomized within task-specific ranges. Paired fine-tuning uses 250 calibration episodes, with 50 episodes for each of five objects differing in size, shape, stiffness, and grasp type. For each paired calibration episode, the same object is rigidly fixed at one tabletop pose: the human grasps it, and the robot subsequently replays the motion through $f_{\theta^0}$. The paired encoder traces supervise mapping refinement through Eq.~\ref{eq:loss}; policy learning for P$^{-}$ and P$^{+}$ uses only human-side demonstrations.

\paragraph{Evaluation protocol.} We train ACT~\cite{act}, Diffusion Policy (DP)~\cite{dp}, and $\pi_{0.5}$~\cite{pi05} under each condition and evaluate each task--backbone--condition combination over 20 autonomous robot rollouts, also with randomized initial object positions. We report the percentage of rollouts that complete the task's main action sequence. For Air Freshener Spray, success requires a secure grasp and a clearly initiated downward press; full actuator depression and spray emission are not required because of the hand's actuation-force limit. The collection-efficiency comparison uses successful AirPods demonstrations collected by the same operator within a 30-minute window for each collection method, excluding the one-time paired-replay and optimization overhead.

\begin{table*}[t]
    \centering
    \small
    \setlength{\tabcolsep}{3.9pt}
    \renewcommand{\arraystretch}{1.15}
    \begin{tabular}{l ccc ccc ccc ccc ccc}
    \toprule
    & \multicolumn{3}{c}{\textbf{Screw Driving}} & \multicolumn{3}{c}{\textbf{AirPods Case}} & \multicolumn{3}{c}{\textbf{Ball Throwing}} & \multicolumn{3}{c}{\textbf{Table Cleaning}} & \multicolumn{3}{c}{\textbf{Air Freshener}} \\
    \cmidrule(lr){2-4}\cmidrule(lr){5-7}\cmidrule(lr){8-10}\cmidrule(lr){11-13}\cmidrule(lr){14-16}
    \textbf{Policy} & T & P$^{-}$ & P$^{+}$ & T & P$^{-}$ & P$^{+}$ & T & P$^{-}$ & P$^{+}$ & T & P$^{-}$ & P$^{+}$ & T & P$^{-}$ & P$^{+}$ \\
    \midrule
    ACT~\cite{act}      & 80.0 & 55.0 & 70.0 & 65.0 & 65.0 & 75.0 & 60.0 & 60.0 & 70.0 & 75.0 & 50.0 & 65.0 & \textbf{90.0} & 70.0 & 85.0 \\
    DP~\cite{dp}        & 65.0 & 40.0 & 55.0 & 55.0 & 55.0 & 60.0 & 50.0 & 50.0 & 55.0 & 60.0 & 40.0 & 55.0 & 70.0 & 45.0 & 60.0 \\
    $\pi_{0.5}$~\cite{pi05} & \textbf{90.0} & 60.0 & 80.0 & 75.0 & 75.0 & \textbf{85.0} & 70.0 & 70.0 & \textbf{80.0} & \textbf{85.0} & 60.0 & 75.0 & 85.0 & 65.0 & 80.0 \\
    \bottomrule
    \end{tabular}
    \caption{\textbf{Main results.} Success rate (\%) from 20 autonomous robot rollouts per entry. \textbf{T}~=~policies trained on robot-side teleoperation demonstrations; \textbf{P$^{-}$}~=~\method{} without paired fine-tuning; \textbf{P$^{+}$}~=~\method{} with paired fine-tuning.}
    \label{tab:main}
    \vspace{-1.0em}
\end{table*}

\subsection{Key Findings}
\label{sec:eval:findings}

\paragraph{F1: \method{} improves fine and dynamic tasks.} Averaged over all tasks and policy backbones, teleoperation reaches 71.7\% success, while P$^{+}$ reaches 70.0\% and P$^{-}$ reaches 57.3\%. Thus paired fine-tuning preserves nearly all of the robot-side data quality while removing the robot from the human collection loop. The task-level pattern is more informative: P$^{+}$ outperforms T on AirPods Case Insertion (73.3\% vs.\ 65.0\%) and Ball Basket Throwing (68.3\% vs.\ 60.0\%). These are precisely the settings where real-time teleoperation is least natural, because small orientation corrections and fast grasp-to-release timing are hard to execute through the delayed robot control loop.

\paragraph{F2: Paired fine-tuning matters most under contact load.} Comparing P$^{-}$ to P$^{+}$ isolates the contribution of paired-replay fine-tuning. The average lift is +12.7 percentage points, but it is concentrated on contact-load-dominated tasks: Screw Driving improves by +16.7 points, Table Cleaning by +15.0 points, and Air Freshener Spray by +15.0 points. This matches our expectation: the babbling-only mapping $f_{\theta^0}$ is trained in free space and drifts under finger compliance. The paired-replay loss (Eq.~\ref{eq:loss}) corrects precisely this regime.

\paragraph{F3: \method{} is significantly more efficient than teleoperation.} 
\begin{wrapfigure}{r}{0.42\linewidth}
    \vspace{-1.0em}
    \centering
    \includegraphics[width=\linewidth]{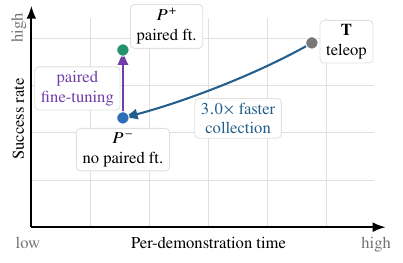}
    \caption{\textbf{Efficiency--quality trade-off.} }
    \label{fig:efficiency}
    \vspace{-1.0em}
\end{wrapfigure}
We compare data collection throughput on the AirPods Case Insertion task within a fixed 30-minute window. The same operator collected 18 successful demonstrations via teleoperation and 52 via \method{}, an almost $3.0\times$ throughput gain. Because P$^{+}$ and teleoperation have nearly matched overall rollout success (70.0\% vs.\ 71.7\%), the success-normalized gain in \emph{useful training data per minute} remains approximately $2.9\times$ (Fig.~\ref{fig:efficiency}). The practical benefit of \method{} is therefore not that it sacrifices data quality for speed, but that it decouples human demonstration from real-time robot execution while paired fine-tuning recovers most of the quality gap.

\section{Conclusions}
\label{sec:conclusion}

We present \method{}, a scalable and efficient demonstration collection and policy learning framework that uses a shared exoskeleton as the interface between human demonstrations and robot execution. By letting the human operator and robot hand wear the same mechanism, \method{} transfers natural contact-rich human motion into robot actions through paired encoder supervision while preserving aligned raw wrist observations. Through challenging real-world experiments, we demonstrate \method{}'s capability in learning precise, contact-rich, and dynamic dexterous manipulation policies, reaching 70.0\% mean success while collecting data $3.0\times$ faster than teleoperation. Our work establishes a new approach to collecting real-world dexterous hand data efficiently and at scale beyond traditional teleoperation.


\section{Limitations}
\label{sec:limitation}

  Although \method{} improves the efficiency of dexterous demonstration collection, several limitations remain. The current system is not yet fully automated across dexterous hands:
  adapting to different link layouts, joint limits, and actuation ranges still requires engineering choices in exoskeleton geometry, calibration motions, and paired regression targets.
  Because operators wear the exoskeleton during collection, wearability, donning time, and task convenience also affect long-horizon and delicate manipulation. Future work will automate
  cross-hand design and mapping, improve the wearing experience, and incorporate tactile sensing for more accurate contact estimation.
\clearpage

\bibliography{main}  

\end{document}